\documentclass[conference]{IEEEtran}
\IEEEoverridecommandlockouts

\usepackage{cite}
\usepackage{amsmath,amssymb,amsfonts}
\usepackage{algorithmic}
\usepackage{graphicx}
\usepackage{textcomp}
\usepackage{xcolor}

\def\BibTeX{{\rm B\kern-.05em{\sc i\kern-.025em b}\kern-.08em
    T\kern-.1667em\lower.7ex\hbox{E}\kern-.125emX}}

\newcounter{MYtempeqncnt}
\newcommand\norm[1]{\left\lVert#1\right\rVert}

\title{Solving Inverse Problems with Score-Based Generative Priors learned from Noisy Data \\
\thanks{This work was supported by NSF IFML 2019844, NSF CCF-2239687 (CAREER), and a gift made by InterDigital, an affiliate of the 6G@UT center within the Wireless Networking and Communications Group at The University of Texas at Austin.}
}

\makeatletter
\newcommand{\linebreakand}{%
  \end{@IEEEauthorhalign}
  \hfill\mbox{}\par
  \mbox{}\hfill\begin{@IEEEauthorhalign}
}
\makeatother

\author{Asad Aali$^1$, Marius Arvinte$^{1,2}$, Sidharth Kumar$^1$, Jonathan I. Tamir$^1$ \\
\IEEEauthorblockA{\textit{$^1$Chandra Family Department of Electrical and Computer Engineering, The University of Texas at Austin}, Austin, TX, USA \\
\textit{$^2$Intel Corporation}, Hillsboro, OR, USA \\
Email: \{asad.aali, arvinte, sidharth.kumar, jtamir\}@utexas.edu}}

\begin{document}
\maketitle

\begin{abstract}
We present SURE-Score: an approach for learning score-based generative models using training samples corrupted by additive Gaussian noise. When a large training set of clean samples is available, solving inverse problems via score-based (diffusion) generative models trained on the underlying fully-sampled data distribution has recently been shown to outperform end-to-end supervised deep learning. In practice, such a large collection of training data may be prohibitively expensive to acquire in the first place. In this work, we present an approach for approximately learning a score-based generative model of the clean distribution, from  noisy training data. We formulate and justify a novel loss function that leverages Stein's unbiased risk estimate to jointly denoise the data and learn the score function via denoising score matching, while using only the noisy samples. We demonstrate the generality of SURE-Score by learning priors and applying posterior sampling to ill-posed inverse problems in two practical applications from different domains: compressive wireless multiple-input multiple-output channel estimation and accelerated 2D multi-coil magnetic resonance imaging reconstruction, where we demonstrate competitive reconstruction performance when learning at signal-to-noise ratio values of $0$ and $10$ dB, respectively.
\end{abstract}


\section{Introduction}
\label{sec:intro}
Recent advances in score-based (diffusion) generative modeling \cite{song2019generative,ho2020denoising} have helped substantially improve the capabilities of solving ill-posed imaging inverse problems using fewer measurements and with higher reconstruction fidelity in various domains such as medical imaging \cite{jalal2021robust,song2021solving,chung2022score,luo2023bayesian}, digital communications \cite{arvinte2022mimo,zilberstein2022annealed}, image super-resolution \cite{kawar2022denoising}, and more.

However, learning high-quality score-based generative models for distributions over real-world signals currently assumes a large database of fully-sampled and noise-free training samples are available \cite{koehler2022statistical}. In many application domains, acquiring such a training set is impossible in practice, because noise is inherently present in the sensors used to acquire measurements. For example, noise is always present in single or multi-coil magnetic resonance imaging (MRI) due to thermal noise in the radio-frequency (RF) hardware as well as losses due to the body. Similarly, noise is present in communication transceivers, especially in wireless scenarios where mobile users operate devices under cost and energy constraints.
Learning score-based generative models using only noisy samples is thus an important research problem.
In this work, we propose a solution to this problem by combining denoising using Stein's unbiased risk estimate (SURE) \cite{stein1981estimation} with the denoising score matching (DSM) objective used in diffusion model training \cite{vincent2011connection}, to achieve joint denoising and score learning with a single deep learning model.

Our framework connects the two already-similar objectives by leveraging Tweedie's single-step denoising formula \cite{efron2011tweedie} at a properly chosen noise level. We show that this objective is naturally compatible with the DSM formulation and allows for a simple modification which introduces two additional terms and a scalar weight in the training function, in the case of training data corrupted by independent and identically distributed (i.i.d.) Gaussian noise. We evaluate the effectiveness of our proposed approach by using the learned models as priors in two different applications: compressive multiple-input multiple-output (MIMO) channel estimation and accelerated MRI reconstruction. Our results show that the priors learned using the proposed formulation can be reliably used for solving linear inverse problems, even when the training dataset is captured in a poor signal-to-noise ratio (SNR) of $0$ dB.

\subsection{Related Work}
There is an extensive body of prior work related to denoising of structured signals; see \cite{elad2023image} for a recent review paper. Most relevant to our work is the use of the self-supervised SURE objective \cite{stein1981estimation}, which introduces an unbiased estimate of the supervised denoising loss purely using noisy data. Using the SURE loss as a training objective for deep neural network denoisers has been previously investigated in a standalone denoising setting \cite{soltanayev2018training,metzler2018unsupervised,kim2021noise2score}, and the connection to multi-level denoising score matching has only been recently explored in \cite{arvinte2022score}, where learning the score function on a restricted subset of noise levels is addressed.

Several other recent methods have used self-supervised learning for blind denoising without ground truth training data. One example is Noise2Noise \cite{lehtinen2018noise2noise}, where multiple noisy measurements of the same sample are used to train a deep neural denoiser. The work in \cite{batson2019noise2self} further introduces the Noise2Self framework, which requires only a single noisy measurement of each sample in the training set, similar to our proposed method. The recent work in \cite{kim2021noise2score} introduces the Noise2Score framework for self-supervised learning of the score function in moderate and high SNR settings, which leaves the problem of learning score functions in low SNR settings open. More recently, the work in \cite{xiang2023ddm} has investigated learning a score-based generative model from noisy diffusion MRI samples, still requiring access to multiple noisy scans of the same subject.

Pre-trained generative priors have also been used in solving linear inverse problems, surpassing classical compressed sensing approaches \cite{lustig2008compressed,alkhateeb2015compressed}. Recently, a series of works have shown that score-based generative models produce competitive reconstruction performance when used to solve accelerated MRI \cite{jalal2021robust,song2021solving,chung2022score,luo2023bayesian} in a supervised learning setting, and that the score can be learned implicitly from MMSE denoisers \cite{kadkhodaie_prior_implicit_denoiser}. The work in \cite{arvinte2022mimo} applies the same ideas to learning priors for MIMO channel estimation from a limited number of pilot measurements. The recent work of \cite{cui2022self} explores self-supervised learning of score models for MRI reconstruction by splitting the measurement data into distinct subsets, similar in approach to Noise2Self.

\subsection{Contributions}
Our contributions in this work are the following:
\begin{itemize}
    \item We formulate SURE-Score: a framework for jointly learning a denoiser and score-based generative model using the same deep neural network. The loss connects SURE, DSM, and Tweedie's rule in a single objective.
    \item We evaluate SURE-Score by learning score-based generative models from noisy data for MIMO channels and MRI. Even though the distributional structure in the two domains is different, our results show that SURE-Score can learn accurate priors even at an SNR of $0$ dB.
    \item Our simulation results show that MIMO channels acquired at $0$ dB can be used for compressed channel estimation with a performance drop of at most $8$ dB in normalized mean square error (NMSE) compared to fully supervised. We also show that self-supervised denoising and reconstruction at $5\times$ accelerated MRI is possible with about $15\%$ error increase when learning a prior from noisy samples acquired at $10$ dB.
\end{itemize}

We use lowercase and uppercase letters to denote vectors (including scalars) and matrices, respectively. We use $p_X$ to denote the probability distribution function (p.d.f.) of the random variable $X$. We use $\underbar{x}$ to denote the flattened version of $X$, and $X^\mathrm{H}$ to denote the Hermitian of $X$. Unless explicitly stated, all signals are assumed to be complex-valued.

\section{System Model and Background}
\label{sec:system_model}
\subsection{Linear Inverse Problems}
We model signals corrupted by additive white Gaussian noise as follows:
\begin{equation}
\label{eq:noisy_samples}
    \tilde{x} = x + w,
\end{equation}
\noindent where $x \in \mathbb{C}^{N}$  is sampled from the distribution $p_X$. The noise $w$ $\sim \mathcal{CN}(0,\sigma^{2}_wI)$ is i.i.d. complex-valued Gaussian distribution, $I$ is the identity matrix, and $\sigma^2_w$ is the noise power. In the general case of linear inverse problems, the measurements are given by:
\begin{equation}
\label{eq:inverse_problem}
    y = Ax + n,
\end{equation}
\noindent where $x \sim p_X$ is the ground-truth signal, $y\in\mathbb{C}^M$ are the observed measurements, $n \sim \mathcal{CN}(0, \sigma_n^2 I)$ is the measurement noise, and $A\in\mathbb{C}^{M\times N}$ is the forward operator. The goal of solving linear inverse problems is to estimate the signal $x$, given the measurements $y$, the forward operator $A$, and statistical knowledge about the distributions of $x$ and $n$.

\subsection{Score-Based Generative Modeling}
\label{subsec:dsm}
The score function $\psi_X$ of a probability distribution $p_X$ is defined as the gradient of its logarithm with respect to the sample \cite{hyvarinen2005estimation}:
\begin{equation}
    \psi_X (x) = \nabla_x \log p_X (x).
\end{equation}
Learning the score function of the probability distribution avoids the challenges of normalization and is useful to downstream tasks such as sampling from the learned distribution or solving inverse problems \cite{jalal2021robust}. The goal of score-based generative modeling is to learn the score function $\psi_X$, given a training dataset sampled from $p_X$.

Denoising score matching \cite{vincent2011connection} learns the score of a family of perturbed signal distributions through a noise-conditional score network $s_\theta(x; \sigma)$. When conditioned on a noise level $\sigma$, the network is trained to predict the score of the distribution $p_{\tilde{X}}$ of perturbed real-valued signals $\tilde{x} = x + z$, where $z \sim \mathcal{N}(0, \sigma^2 I)$. The DSM loss is given by:
\begin{equation}
    L_{\textrm{DSM}, \theta}(x; \sigma) = \left\lVert s_\theta (\tilde{x}; \sigma) - \nabla_{\tilde{x}} \log p_{\tilde{X} | X}(\tilde{x} | x; \sigma) \right\rVert^2_2.
\end{equation}
When noiseless data are available for training and choosing the added noise $z$ as Gaussian i.i.d., the gradient of the conditional distribution can be expressed in closed form as $-z/\sigma^2$. Score-based generative models are trained using denoising score matching at multiple noise levels simultaneously \cite{song2019generative}. Following \cite{song2020improved}, we weigh the loss at each level by $\sigma^2$ to normalize the magnitude of the DSM loss across all noise levels. This yields the expression for the multilevel DSM training loss:
\begin{equation}
\label{eq:dsm_multilevel}
L_{\textrm{DSM}, \theta} = \mathbb{E}_{\substack{x \sim p_X \\ \sigma \sim p_\Sigma \\ z \sim \mathcal{N}(0, \sigma^2I)}} \sigma^2 \left\lVert s_\theta (x + z; \sigma) + \frac {z} {\sigma^2} \right\rVert^2_2.
\end{equation}
The distribution of training noise levels $p_\Sigma$ is a uniform distribution across geometrically distributed noise levels between $\sigma_\textrm{min}$ and $\sigma_\textrm{max}$. We use the guidelines in \cite{song2020improved} and implement the noise-conditional score network as $s_\theta (x; \sigma) = s_\theta / \sigma$, where $s_\theta(x)$ is a deep neural network with learnable weights $\theta$.

\subsection{Solving Inverse Problems using Annealed Langevin Dynamics}
The score-based generative model $s_\theta$ learned during training does not use any knowledge of the linear operator $A$ or received measurements $y$ in the setting of \eqref{eq:inverse_problem}, but is still a powerful tool for solving linear inverse problems.
The algorithm we use to generate a point estimate $x_\mathrm{est}$ is based on posterior sampling, proven in \cite{jalal2021instance} to be sample optimal for linear inverse problems. This is a stochastic algorithm that draws a single sample from the posterior distribution: 
\begin{equation}
    x_\mathrm{est} \sim p_{X | Y} (x | y).
\end{equation}
We use a modified version of annealed Langevin dynamics \cite{song2019generative} to sample from the posterior distribution. This is an iterative algorithm that starts from a random $x_0 \sim N(0, I)$ and requires access to the noise-conditional score function $\psi_{X_t |Y} (x_t |y; \sigma)$ at each step. For brevity, we omit $\sigma_t$ from the notation in the following. The update rule is given by:
\begin{equation}
\label{eq:ald}
    x_{t+1} = x_t + \alpha_t \cdot \psi_{X_t|Y} (x_t | y) + \sqrt{2 \beta \alpha_t} \eta_t,
\end{equation}
where $\alpha_t$ is a time-varying step size, $\eta_t\sim \mathcal{N}(0, I)$ is annealing noise sampled and added at each update step, and $\beta$ is a hyper-parameter controlling the level of annealing noise. When $\beta = 1$, the algorithm in \eqref{eq:ald} becomes annealed Langevin dynamics. Following \cite{song2020improved}, we exponentially decay the step size as $\alpha_t = \alpha_0 \cdot \left(\sigma_t / \sigma_\mathrm{min}\right)^2$, defining the schedule via two hyper-parameters. The term $\sigma_{t}$ comes from the geometric distribution of training noise levels beginning from $\sigma_\mathrm{max}$ and ending at $\sigma_\mathrm{min}$.

The conditional score function $\psi_{X_t|Y}$ can be further expanded via Bayes' rule as:
\begin{equation}
\label{eq:bayes}
\begin{split}    
\psi_{X_t | Y}(x_t | y) & = \nabla_{x_t} \log p_{X_t|Y} (x_t|y) \\ & = \nabla_{x_t} \log \frac{p_{Y|X_t}(y|x_t) p_{X_t}(x_t)}{p_Y(y)} \\ & = \psi_{Y | X_t}(y | x_t) + \psi_{X_t}(x_t),
\end{split}
\end{equation}
\noindent where the denominator term vanishes due to not depending on $x$. The first term is related to the linear formulation of the inverse problem in \eqref{eq:inverse_problem}. As $\sigma_t \rightarrow 0$, this term becomes:
\begin{equation}
    \psi_{Y|X_t}(y|x_t) = \frac{A^\mathrm{H} (y - Ax_t)}{\sigma_n^2},
\end{equation}
\noindent where $A^{\mathrm{H}}$ is the Hermitian transpose (adjoint) of $A$. We approximate the dependency on the iteration-dependent noise level $\sigma_t$ in the first term of \eqref{eq:bayes} by additionally annealing this term with a factor $\gamma_t$
as in \cite{jalal2021robust}:
\begin{equation}
    \psi_{Y|X_t}(y|x_t) \approx \frac{A^\mathrm{H} (y - Ax_t)}{\sigma_n^2 + \gamma_t^2},
\end{equation}
where we choose $\gamma_t^2 = \sigma_t^2$ in this work. The dependency of the second term in \eqref{eq:bayes} on $\sigma_t$ is included in a pre-trained noise-conditional score-based model $s_\theta(x; \sigma_t)$, which can readily approximate this term. The final update rule for sampling from the posterior distribution is given by:
\begin{equation}
x_{t+1} = x_t + \alpha_t \left( \frac{A^\mathrm{H} (y - A x_t)}{ \sigma_n^2 + \gamma_t^2 } + s_\theta (x_t; \sigma_t) \right) + \sqrt{2 \beta \alpha_t } \eta_t.
\label{eqn:our_algorithm}
\end{equation} 
We emphasize that our algorithm only approximates posterior sampling.

\subsection{Stein's Unbiased Risk Estimate}
\label{subsec:sure_loss}
It is clear from Section~\ref{subsec:dsm} that learning a prior via denoising score matching requires a clean dataset of training set $x \sim p_X$. However, in our setting we only have access to a training set of noisy samples $\tilde{x}$ sampled according to \eqref{eq:noisy_samples}.
Ideally, we wish to learn a denoising model $g_\phi (x)$ with parameters $\phi$, such that it maps perturbed samples $\tilde{x}$ to clean samples $x$. One approach is to use Stein's unbiased risk estimate \cite{li1985stein} to obtain an unbiased estimate of the ground-truth supervised denoising loss $\norm{g_\phi(\tilde{x}) - x}_2^2$ for each sample $x$. When the sample is corrupted by i.i.d. Gaussian noise at noise level $\sigma_w$, the SURE expression is given by:
\begin{equation}
    L_{\textrm{SURE}, \phi} (\tilde{x}; \sigma_w) = \norm{\tilde{x} - g_\phi(\tilde{x})}_2^2 + 2 \sigma_w^2 \cdot \mathrm{div}_{\tilde{x}}(g_\phi (\tilde{x})),
\end{equation}
\noindent where $\mathrm{div}_x(g_\phi(x))$ represents the divergence of the vector field (any function that maps inputs to outputs of the same dimension) $g_\phi$ evaluated at the point $x$, and is given by:
\begin{equation}
\label{eq:div_def}
    \mathrm{div}_x(g_\phi(x)) = \sum_i \frac{\partial [g_\phi(x)]_i}{\partial x_i}.
\end{equation}
Evaluating all partial derivatives in \eqref{eq:div_def} requires a number of function evaluations proportional to the input dimension, which does not scale to real-world high-dimensional signals. Following \cite{ramani2008monte}, we approximate the divergence term using its Monte Carlo approximation with $n \sim \mathcal{N}(0, I)$ being i.i.d. Gaussian noise sampled for each sample in each batch, and the approximation given by:
\begin{equation}
\label{eq:div_approx}
\mathrm{div}_x(g_\phi(x)) \approx n^T \left(\frac{g_\phi(x  + \epsilon n) - g_\phi(x )} {\epsilon}\right),
\end{equation}
for small $\epsilon$. The SURE loss does not involve the ground-truth sample $x$ and has the remarkable property that its expectation taken over $\tilde{x} \sim \mathcal{N}(n, \sigma_w^2 I)$ satisfies \cite{stein1981estimation}:
\begin{equation}
    \mathbb{E}_{\tilde{x}}  L_{\textrm{SURE}, \phi} (\tilde{x}; \sigma_w) = \mathbb{E}_{\tilde{x}} \norm{g_\phi(\tilde{x}) - x}_2^2,
\end{equation}
thus it can be used as a training function for the model $g_\theta$, when only a noisy set of training samples is available \cite{ramani2008monte}, and a mini-batch of noisy samples is used. Note that the SURE loss can be applied when a single noisy version of each sample is available in the dataset, which matches our setting.

\begin{figure*}[!t]
\normalsize
\setcounter{MYtempeqncnt}{\value{equation}}
\setcounter{equation}{\value{MYtempeqncnt}}
\stepcounter{MYtempeqncnt}
\stepcounter{MYtempeqncnt}
\begin{align}
L_{\textrm{SURE-Score}, \theta} (\tilde{x}; \sigma, \sigma_w) & = \sigma^2 \norm{s_\theta (g_\theta(\tilde{x}; \sigma_w) + z; \sigma) + \frac{z}{\sigma^2}}_2^2 + \lambda \norm{\tilde{x} - g_\theta (\tilde{x}; \sigma_w)}_2^2 + 2 \lambda \sigma_w^2 \cdot \textrm{div}_{\tilde{x}} (g_\theta (\tilde{x}; \sigma_w)) \label{eq:our_loss_multicol1} \\ & = \sigma^2 \norm{s_\theta (\tilde{x} + \sigma_w^2 s_\theta(\tilde{x}; \sigma_w) + z; \sigma) + \frac{z}{\sigma^2}}_2^2 + \lambda \norm{\sigma_w^2 s_\theta(\tilde{x}; \sigma_w)}_2^2 + 2 \lambda \sigma_w^2 \cdot\textrm{div}_{\tilde{x}} (\tilde{x} + \sigma_w^2 s_\theta(\tilde{x}; \sigma_w)). \label{eq:our_loss_multicol2}
\end{align}
\setcounter{equation}{\value{MYtempeqncnt}}
\hrulefill
\vspace*{4pt}
\end{figure*}

\section{Proposed Method: SURE-Score}
\begin{figure}[t]
\centerline{\includegraphics[width=1\linewidth]{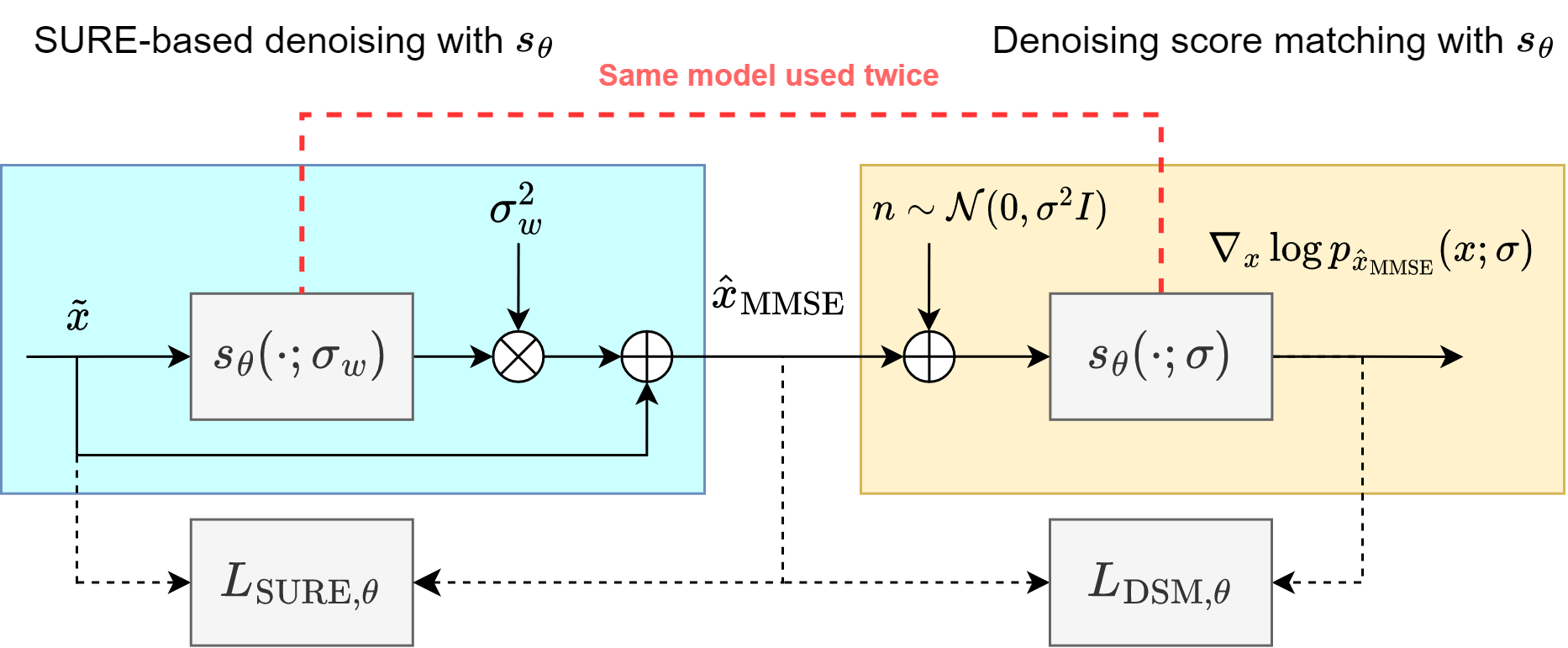}}
\caption{Flow of SURE-Score during training. The same deep neural network $s_\theta$ is used first for denoising and subsequently for denoising score matching. The terms in \eqref{eq:our_loss_multicol2} are obtained from the input, $\hat{x}_\mathrm{MMSE}$, and the final output.}
\label{fig:block_diagram}
\end{figure}

Our proposed method learns the score of the perturbed distribution of minimum mean square error (MMSE) denoised samples. While this is distinct from learning the score of the true perturbed distribution of clean samples at arbitrary noise levels, our formulation couples SURE learning and denoising score matching via Tweedie's rule for Gaussian corruptions such that the two objectives are consistent with each other at a specific noise level.
We call our method SURE-Score. A block diagram of the training flow is shown in Figure~\ref{fig:block_diagram}.

Naively, learning the score of perturbed MMSE denoised samples could be done with two different functions, each with its own set of learnable parameters, and applied sequentially:
\begin{enumerate}
    \item A function $g_\phi$ that learns the MMSE denoiser. This function can be learned purely from noisy data using the SURE training objective described in Section~\ref{subsec:sure_loss}.
    \item A score-based generative model $s_\theta$ trained on the outputs of $g_\phi$. If sufficient training data are available, it could be split between the two stages, otherwise the same training set used to learn $g_\phi$ can be re-used in the second stage.
\end{enumerate}
Inspired by the recent work in \cite{kim2021noise2score}, we take advantage of the inherent connection between the two steps given by Tweedie's rule \cite{efron2011tweedie} which allows us to efficiently re-use the same network parameters for both purposes and reduce storage requirements. Assuming that a score-based generative model $s_\theta(\tilde{x}; \sigma)$ is available and already learned, the MMSE denoiser in the case of AWGN is given by its \textit{Tweedie re-parameterization} as \cite{kim2021noise2score}:
\begin{equation}
    g_\theta(\tilde{x}; \sigma) = \tilde{x} + \sigma^2 \cdot s_\theta(\tilde{x}; \sigma).
    \label{eq:tweedie}
\end{equation}
Conversely, given access to an already trained MMSE denoiser $g_\theta(\tilde{x}; \sigma)$ its score re-parameterization is given by \cite{kim2021noise2score}:
\begin{equation}
    s_\theta(\tilde{x}; \sigma) = \frac{g_\theta(\tilde{x}; \sigma) - \tilde{x}}{\sigma^2},
\end{equation}
\noindent where the above is due to the existence and uniqueness of the MMSE estimator under finite means and variance of the perturbed distribution of $\tilde{x}$, hence Tweedie's rule can be directly inverted for $\sigma > 0$.

The previous identities suggest that a score-based generative model for a specific noise level and the MMSE denoiser for the same noise level are interchangeable. Thus, we propose to learn them \textit{jointly} using the following loss function at a single noise level $\sigma$, with training data corrupted at noise level $\sigma_w$:
\begin{equation}
\label{eq:our_abstract_loss}
\begin{split}
    L_{\textrm{SURE-Score}, \theta}(\tilde{x}; \sigma, \sigma_w) = \ & L_{\textrm{DSM}, \theta} (g_\theta(\tilde{x}; \sigma_w); \sigma) + \\ & \lambda \cdot L_{\textrm{SURE}, \theta}(\tilde{x}; \sigma_w),
\end{split}
\end{equation}
\vspace{-2mm} 

\noindent where $\lambda>0$ is a hyper-parameter controlling the relative weighting between the two losses. Note that in the above, the DSM term uses the output of the denoiser $g_\theta$ -- this leads to a recurrent call of the same underlying network. Each of the two terms in the above corresponds to one of two objectives: the first term learns the score-based generative model for the perturbed distribution of MMSE denoised samples, while the second term learns the MMSE denoiser itself.

The Tweedie re-parameterization introduces a consistency between the two objectives at noise level $\sigma_w$, that is when $\sigma = \sigma_w$ in \eqref{eq:our_abstract_loss}: the more accurately the true score function is approximated at noise level $\sigma_w$, the more accurately denoising 
becomes, and vice versa. This can be viewed as a coarse version of the recently introduced consistency objective in \cite{daras2023consistent}, where we ensure that removing and adding back noise at noise level $\sigma_w$ leads to consistently learned gradients of the perturbed score function. In practice, both terms at $\sigma = \sigma_w$ are bounded and non-zero due to the non-zero residuals of MMSE denoising. Additionally, the above consistency is only introduced for $\sigma = \sigma_w$, but in practice we find that this does not hinder performance at other noise levels, where extrapolation is sufficiently accurate.

Expanding the expected versions of the two losses in \eqref{eq:our_abstract_loss} yields the expression in \eqref{eq:our_loss_multicol1}. Replacing $g_\theta$ with its Tweedie re-parameterization yields our SURE-Score loss function in \eqref{eq:our_loss_multicol2}, which only depends on the score network $s_\theta$ and regularization $\lambda$. The training loss we use in practice is a mini-batch version of \eqref{eq:our_loss_multicol2}, with expectations identical to those in \eqref{eq:dsm_multilevel}.

Note that while all the derivations so far have relied on the assumption that $\tilde{x}$ is corrupted by i.i.d. Gaussian noise, both the SURE loss and the Tweedie re-parameterization can be readily extended to other exponential family distributions \cite{eldar2008generalized,kim2021noise2score}, making our proposed approach easily applicable to other common noise models as well.

Finally, our training and reconstruction algorithms require the selection of several hyperparameters without access to ground-truth for tuning. In particular, we inherit all hyperparameters related to both denoising score matching and to SURE; notable among them is the step size schedule in DSM given by $p_\Sigma$, which is known to depend on the dataset \cite{song2020improved}, and the Monte Carlo approximation $\epsilon$, which is known to depend on the noise level $\sigma_w$ \cite{ramani2008monte,arvinte2022score}. We also must choose $\lambda$ to balance the two losses in \eqref{eq:our_loss_multicol2}. 

\section{Experimental Results and Discussion}
We test our approach on two linear inverse problems: estimating the MIMO wireless channel matrix from compressed pilot measurements, and multi-coil MRI reconstruction from limited k-space measurements. We compare our approach to following baselines: (i) supervised score-model training, (ii) naive DSM training on the noisy data, (iii) BM3D, (iv) Noise2Self, and (v) Noise2Score. In (iii-iv), we apply DSM training on top of the denoised dataset.

We qualitatively evaluate score models learned from training data at different SNR levels (given by SNR$^w$). We also quantitatively evaluate posterior reconstruction performance. Following convention, we report normalized mean squared error (NMSE) in dB for channel estimation and normalized root mean squared error (NRMSE) for MRI, each at particular sub-sampling ratios and for different values of $\sigma_w^2$ and $\sigma_n^2$.

\subsection{Model Architecture}
\label{subsec:model_arch}
We use a deep neural network based on the RefineNet model \cite{lin2017refinenet}, following the NCSNv2 architecture in \cite{song2020improved}\footnote{Exact implementation will be made available in our source code upon paper publication.}.
To handle complex-valued signals for both applications, we treat the real and imaginary parts as two real-valued channels of the same sample.
As suggested by other works \cite{metzler2018unsupervised}, we set $\epsilon = 10^{-3}$ in \eqref{eq:div_approx}. We use the Adam optimizer with a fixed learning rate of $10^{-4}$ for all experiments. The parameter $\lambda$ in \eqref{eq:our_abstract_loss} was set to $L_{\mathrm{DSM}, \theta} / L_{\mathrm{SURE}, \theta}$ using the first batch at the start of training.

\subsection{MIMO Channel Modeling and Estimation}
We consider a point-to-point MIMO baseband communication scenario where the transmitter and receiver are equipped with $N_\mathrm{t}$ and $N_\mathrm{r}$ antennas, respectively. Channel estimation involves estimating the channel matrix $H \in \mathbb{C}^{N_\mathrm{r} \times N_\mathrm{t}}$, given measurements $Y$ of the form:
\begin{equation}
\label{eq:mimo_system_matrix}
    Y = H P + N,
\end{equation}
\noindent where the matrix $P \in \mathbb{C}^{N_\mathrm{t} \times N_\mathrm{p}}$ is the measurement matrix known ahead of time (commonly called a pilot matrix), and $N$ is additive white Gaussian noise with variance $\sigma_{\textrm{n}}^2$. Equation~\eqref{eq:mimo_system_matrix} can be vectorized as:
\begin{equation}
\label{eq:mimo_system_vector}
    \underbar{y} = \left( P^\mathrm{H} \otimes I_{N_\mathrm{r}} \right) \underbar{h} + \underbar{n},
\end{equation}
\noindent where $I_{N_\mathrm{r}}$ represents the square identity matrix of size $N_\mathrm{r}$. Thus, this problem falls in the general class of linear, noisy inverse problems given by \eqref{eq:inverse_problem}. We use randomly chosen quadrature phase shift keying (QPSK) signals for the entries of $P$. Channel estimation is an under-determined inverse problem when the pilot density ratio $\alpha = N_\mathrm{p} / N_\mathrm{t} < 1$. We normalize channels during training and inference by dividing by $\mathbb{E}_{H_\mathrm{train}}[\norm{H}_2^2]$. The SNR for a distribution of channels is defined as: $\mathbb{E}_H [\norm{H}_2^2] / \sigma^2_w = 1/\sigma_w^2$.

We train a score-based model $s_\theta$ with $32$ hidden layers in the first block using the proposed SURE-Score framework on perturbed clustered delay line (CDL) channels $\tilde{H}$ at SNR$^w = 0$ dB, in particular from the CDL-C non-line-of-sight distribution. We simulate $10000$ (possibly noisy) samples for training for each method where $N_\mathrm{r}=16$ and  $N_\mathrm{t}=64$.

\begin{figure}[ht]
\includegraphics[width=0.9\linewidth]{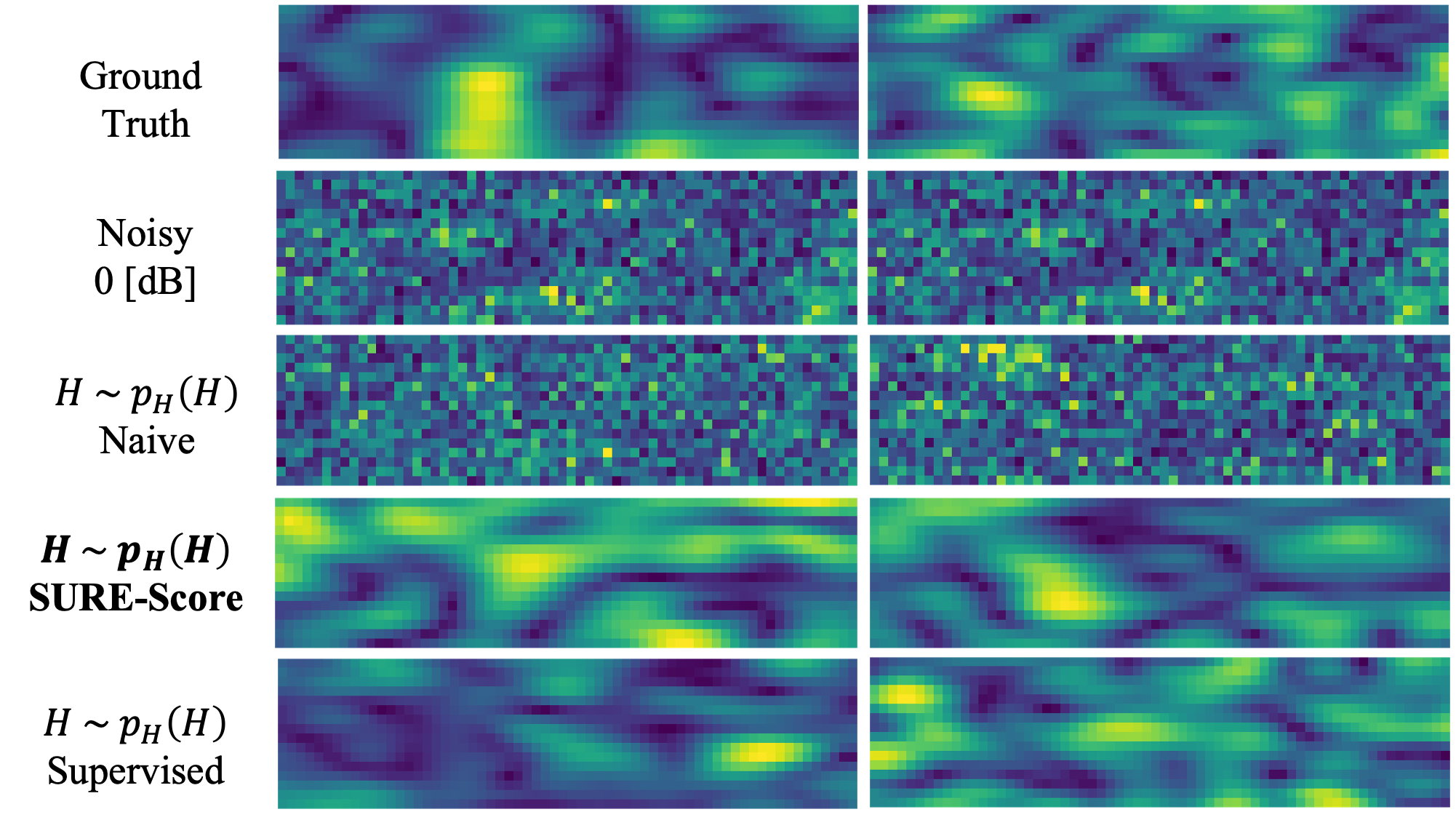}
\caption{Prior sampling performance for three methods: naive (ii), SURE-Score at SNR$^w = 0$ dB, and supervised (i). Each column shows a different realization of CDL-C channels.} 
\label{fig:prior_mimo}
\end{figure}
\subsubsection{Prior Sampling}
To sample unconditionally from the learned prior $p_{\hat{x}_{\mathrm{MMSE}}}$, we use the update in \eqref{eqn:our_algorithm} with
$A=0$, eliminating the measurement-dependent term.
Figure~\ref{fig:prior_mimo} shows CDL-C samples from the training set, noisy samples at SNR$^w$, and samples using three learned priors: (i) naive DSM, (ii) SURE-Score, and (iii) DSM with clean data. Qualitatively, we can see that prior samples generated from a score model naively trained on noisy channels are corrupted. However, both supervised and SURE-Score samples qualitatively match the signal structure in the reference samples. This is further validated in our posterior sampling experiments.

\begin{figure}[ht]
\begin{center}
\centerline{\includegraphics[width=0.9\linewidth]{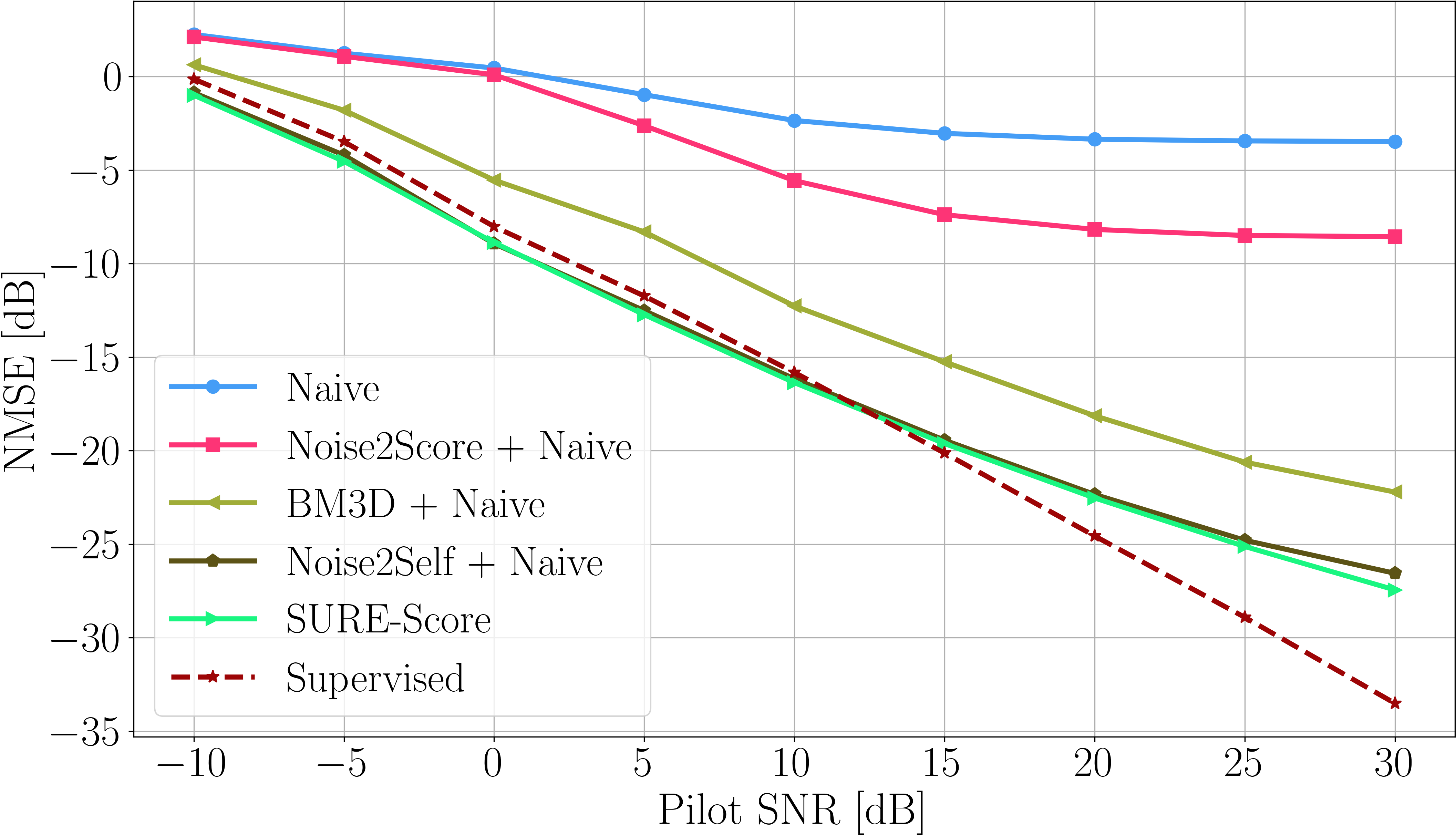}}
\vspace{2mm} 
\centerline{\includegraphics[width=0.9\linewidth]{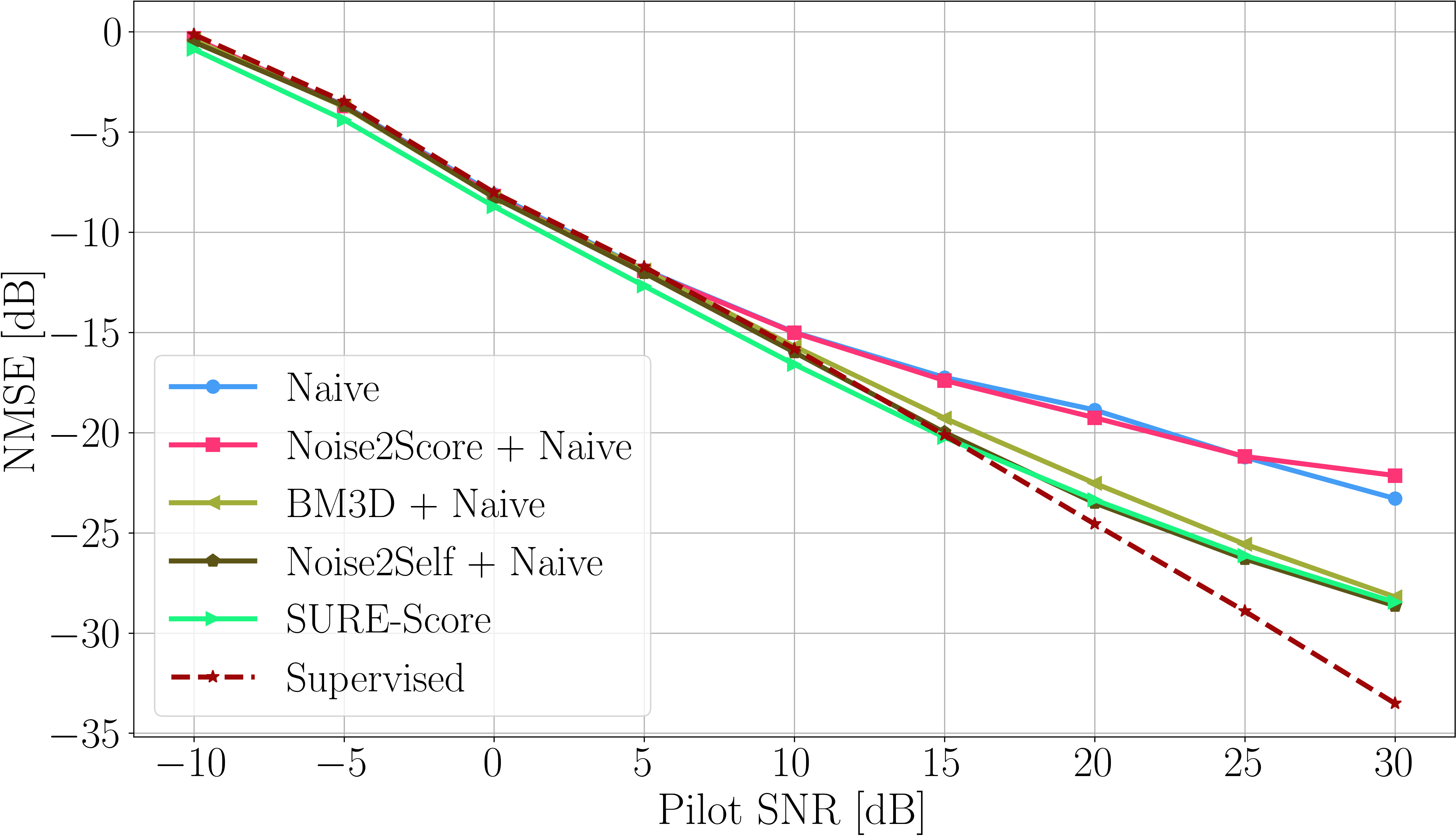}}
\end{center}
\vspace{-6mm} 
\caption{Channel estimation performance at $\alpha = 0.6$ ($38$ pilots) using score models trained on CDL-C channels at SNR$^w$: $0$ dB (top) and $10$ dB (bottom).}
\label{fig:mimo_posterior_results}
\end{figure}
\subsubsection{Posterior Reconstruction}
\begin{figure*}[t]
\begin{center}
\includegraphics[width=.9\textwidth]{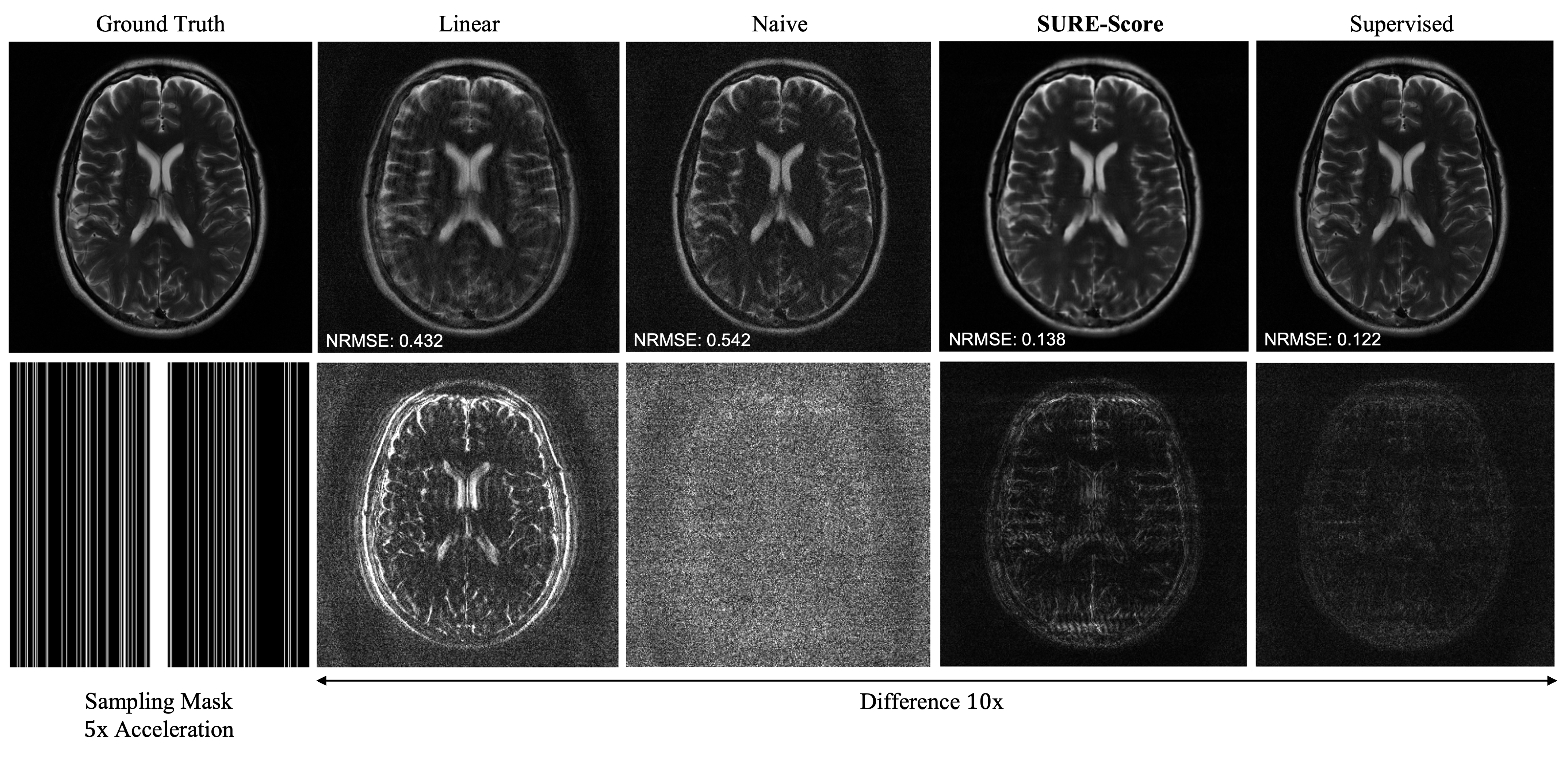}
\end{center}
\caption{Multi-coil MRI reconstruction at acceleration factor of $5\times$. From left to right: fully sampled ground truth, linear reconstruction, posterior sampling after naively training on noisy data at $\mathrm{SNR}^w$, posterior sampling after training with SURE-Score, and posterior sampling after training with noise-free data. The bottom row shows the sampling pattern and difference images for each method, respectively.}
\label{fig:mri_recon}
\end{figure*}
We use the annealed Langevin dynamics update in \eqref{eqn:our_algorithm} with the data consistency term based on \eqref{eq:mimo_system_vector}. We simulate a pilot density of $\alpha=0.6$, i.e., $38$ pilots. Figure~\ref{fig:mimo_posterior_results} summarizes results from posterior sampling using different training methods and at different pilot SNR levels, where we assume training samples $\tilde{H}$ contain noise with SNR of $0$ dB and $10$ dB in the two plots. We observe that naive training plateaus in estimation performance because of overfitting to the noisy training set. Noise2Score and BM3D suffer at lower $\mathrm{SNR}^w$ and improve at higher SNR, consistent with results reported in the literature \cite{kim2021noise2score}. SURE-Score performs close to optimal with respect to supervised DSM except at higher pilot SNR. Noise2Self as a denoising pre-processing step also performed well. We observe a performance gap at high pilot SNR, likely due to the performance limits of the learned MMSE denoiser for our particular architecture and the finite amount of training data.

\subsection{Multi-Coil MRI Reconstruction}
Multi-coil MRI data are acquired in the frequency domain by placing multiple RF coils around the imaging anatomy. We assume a vectorized image $ x \in \mathbb{C}^N$, with $N_\mathrm{c}$ RF receive sensitivity profiles each represented by a diagonal matrix $S_i\in\mathbb{C}^{N\times N}$, and Fourier sampling $F_\alpha\in\mathbb{C}^{\alpha N \times N}$. The acquired k-space data from the $i^{\mathrm{th}}$ coil is given by:
\begin{equation}
    y _i = F_\alpha S_i x + n_i,
\end{equation}
\noindent where $n_i$ is additive white Gaussian noise with variance $\sigma_w^2$.
Following convention, we define the acceleration factor $1/\alpha$ relative to a single fully sampled coil, such that the total number of acquired measurements is $\alpha \cdot N_\mathrm{c} \cdot N$. We can represent the MRI multi-coil forward model according to \eqref{eq:inverse_problem}.

We train score-based models with $128$ hidden layers in the first block using $10000$ multi-coil T2-weighted brain MRI scans from the FastMRI dataset \cite{zbontar2018fastmri}. The SURE-Score model was trained on a subset of the data ($2000$ samples) due to training time constraints. We first normalize all slices in the training set by performing a root sum-of-squares (RSS) reconstruction using k-space data from a central window of size $24\times 24$. We divide the k-space of each slice by the $95^\mathrm{th}$ magnitude percentile of this low-resolution reconstruction. We define the SNR as $1/\sigma_w^2$.

\subsubsection{Denoising Performance}
We evaluate denoising performance using the Tweedie re-parameterization in \eqref{eq:tweedie} for score models trained (i) naively, (ii) with SURE-Score, and (iii) using noise-free data (supervised). Table~\ref{table:mri_denoising} lists the NRMSE (mean and standard deviation) over $100$ validation slices for $\mathrm{SNR}^w$ levels of $0$ dB and $10$ dB. The denoising performance of SURE-Score nearly matches supervised learning, indicating consistency at $\sigma_w$ between the SURE objective and the score-matching objective. Thus, denoising score matching at multiple levels does not degrade denoising performance at $\sigma_w$.

\begin{table}[h]
\centering
\begin{tabular}{|c||c|c|}
\hline
\multicolumn{3}{|c|}{Denoising Performance (NRMSE)} \\
\hline
SNR$^w$ & $0$ dB & $10$ dB \\
\hline
Naive & $2.48 \pm 0.24$ & $0.70 \pm 0.07$ \\
Supervised & $0.21 \pm 0.01$ & $0.14 \pm 0.01$ \\
SURE-Score & $0.23 \pm 0.01 $ & $0.16 \pm 0.01$ \\
\hline
\end{tabular}
\vspace{0.7mm}
\caption{MRI denoising performance of score models trained with noise-corrupted data and different objective functions.}
\vspace{-4mm}
\label{table:mri_denoising}
\end{table}

\subsubsection{Posterior Reconstruction}
We use annealed Langevin dynamics as previously described in \eqref{eq:ald}. We simulate an acceleration factor of five using a vertical sampling mask with fully sampled central k-space and uniform random sub-sampling elsewhere. We assume high SNR and set $\sigma_n=0$ during inference. 

Table~\ref{table:mri_recon} displays reconstruction NRMSE (mean and standard deviation) over a validation set of $50$ slices. We compare linear reconstruction, naive training, supervised training, and SURE-Score. We observe similar trends as in our MIMO experiments. Figure~\ref{fig:mri_recon} compares example reconstructions after training with noisy data at $\mathrm{SNR}^w$. Naive training leads to noise amplification, while SURE-Score qualitatively and quantitatively improves over this result.

\begin{table}[h]
\centering
\begin{tabular}{ |c||c|c|}
 \hline
 \multicolumn{3}{|c|}{Reconstruction Performance (NRMSE)} \\
 \hline
  SNR$^w$ & $0$ dB & $10$ dB \\
 \hline
 Linear & $0.53 \pm 0.19$ & $0.53 \pm 0.19$ \\
 Naive & $1.42 \pm 0.35$ & $0.71 \pm 0.15$ \\
 Supervised & $0.16 \pm 0.05$ & $0.16 \pm 0.05$\\
 SURE-Score & $0.22 \pm 0.06$ & $0.19\pm0.05$ \\
 \hline
\end{tabular}
\vspace{0.7mm}
\caption{Comparison of reconstruction NRMSE (mean and standard deviation) over the validation set for FastMRI.}
\vspace{-4mm}
\label{table:mri_recon}
\end{table}

\section{Conclusion}
Through our results, we can observe the impact of SURE-based denoising on prior sampling as well as for solving inverse problems in the wireless and MR domain while learning from datasets corrupted with a noise-to-signal ratio of up to $0$ dB. We also observe that self-supervised techniques like SURE-Score can match supervised denoising performance. While our algorithm matches the convergence speed of supervised score models in terms of epochs, we find the runtime per iteration increases due to an additional pass through the network at each step. Another challenge, like in other self-supervised methods, is  choosing hyper-parameters without access to ground truth data. Finally, our work currently assumes white Gaussian noise corruption but could be extended to arbitrary exponential families using Generalized SURE \cite{eldar2008generalized}.

\bibliographystyle{IEEEtran}
\bibliography{main} 

\end{document}